\documentclass[conference]{IEEEtran}
\IEEEoverridecommandlockouts

\usepackage{cite}
\usepackage{amsmath,amssymb,amsfonts}
\usepackage[ruled,linesnumbered]{algorithm2e}

\SetCommentSty{mycommfont}
\usepackage{booktabs}
\usepackage{graphicx}
\usepackage{multirow}
\usepackage{hyperref}
\usepackage{textcomp}
\usepackage{xcolor}
\def\BibTeX{{\rm B\kern-.05em{\sc i\kern-.025em b}\kern-.08em
    T\kern-.1667em\lower.7ex\hbox{E}\kern-.125emX}}

\newcommand{\tocheck}[1]{\textcolor{blue}{#1}}
\begin{document}

\title{An Adjustable Farthest Point Sampling Method for Approximately-sorted Point Cloud Data}
\author{\IEEEauthorblockN{Jingtao Li\IEEEauthorrefmark{1}, Jian Zhou\IEEEauthorrefmark{2}, Yan Xiong\IEEEauthorrefmark{1}, Xing Chen\IEEEauthorrefmark{1}, Chaitali Chakrabarti\IEEEauthorrefmark{1}}
\IEEEauthorblockA{\IEEEauthorrefmark{1}School of Electrical Computer and Energy Engineering,
Arizona State University, Tempe, AZ, 85287}
\IEEEauthorblockA{\IEEEauthorrefmark{2}ASML Holding}
\IEEEauthorrefmark{1}\{jingtao1, yxiong35, xchen382, chaitali\}@asu.edu; \IEEEauthorrefmark{2}\{zhoujian1991\}@gmail.com
}
\maketitle
\begin{abstract}
Sampling is an essential part of raw point cloud data processing such as in the popular PointNet++ scheme. Farthest Point Sampling (FPS), which iteratively samples the farthest point and performs distance updating, is one of the most popular sampling schemes. Unfortunately it suffers from low efficiency and can become the bottleneck of point cloud applications.
We propose adjustable FPS (AFPS), parameterized by $M$, to aggressively reduce the complexity of FPS without compromising on the sampling performance. Specifically, it divides the original point cloud into $M$ small point clouds and samples $M$ points simultaneously. It exploits the dimensional locality of an approximately sorted point cloud data to minimize its performance degradation. AFPS method can achieve 22 to 30\texttimes\: speedup over original FPS. Furthermore, we propose the nearest-point-distance-updating (NPDU) method to limit the number of distance updates to a constant number.
The combined NPDU on AFPS method can achieve a 34-280\texttimes\: speedup on a point cloud with 2K-32K points with algorithmic performance that is comparabe to the original FPS. For instance,
for the ShapeNet part segmentation task, it achieves 0.8490 instance average mIoU (mean Intersection of Union), which is only 0.0035 drop compared to the original FPS.
\end{abstract}


\section{Introduction}
With the wide application of LiDAR (Light Detection and Ranging) technology in autonomous driving and self-navigation robots, point cloud data has become a popular data source. However, processing large volume of point cloud data efficiently is a difficult problem especially when it is generated at a fast rate as in LiDAR. 
Recently, PointNet++ \cite{qi2017pointnet++} is shown to achieve excellent performance in processing raw point cloud data with low latency compared to traditional processing schemes that depend on converting point clouds into other formats. 
PointNet++ is based on hierarchical set abstraction layer that consists of sampling and grouping operations, where sampling derives representative points and grouping derives groups of neighboring points of each sampled point.

There are several popular sampling schemes. Of these, Random Point Sampling (RPS) achieves low latency but has poor performance for a sparse point cloud since it can miss out sampling in sparse regions.
Farthest Point Sampling (FPS), which iteratively samples the farthest point and performs distance updating, has been widely accepted \cite{qi2017pointnet++, shi2019pointrcnn, guo2021pct}. However, FPS suffers from poor hardware efficiency and is the latency bottleneck of PointNet++ model when executed on a GPU.

To improve the sampling efficiency, alternative approaches have been investigated. In \cite{lang2020samplenet, hu2020randla}, learning-based methods are used,  \cite{xu2020grid} proposes an efficient and effective Grid-GCN sampling method. The above approaches either requires re-design of the network or an expensive pre-processing step, leading to extra design effort and implementation overhead.

The computational complexity of FPS is O(NC), where $N$ is the number of points in the point cloud and $C$ is the number of sampled points. One way of reducing the time complexity is by dividing the original point cloud data into multiple sectors and performing local FPS on each one of them. The sampling performance of such a method can be quite poor if the original (unsorted) point cloud data is divided arbitrarily. However, we observe performance degradation can be avoided if the point cloud data is sorted or approximately sorted.
In this paper we propose adjustable FPS (AFPS) which divides the original point cloud into multiple point sectors along the (sorted) dimension and performing local FPS on each of them. AFPS exploits the  locality of data in the sorted dimension and is able to maintain algorithmic performance comparable to FPS while achieving significant reduction in the complexity.
Moreover, we propose the nearest-point-distance-updating (NPDU) heuristic to further improve the efficiency of performing local FPS in each small sector.  By only performing distance updates for points close to the newly sampled one, the number of distance update operations in each iteration is reduced from O(N) to a constant number.
With application of NPDU, the AFPS achieves even better efficiency with minimal performance loss in certain cases. Our evaluation of three down-streaming tasks shows the massive speedup of proposed NPDU-AFPS with minor degradation in task performance.
In summary, we make several contributions:
\begin{itemize}
    \item We  propose AFPS, a parametrizable sampling scheme that partitions the point cloud into $M$ point sectors and performs small-scale FPS locally. We show if the point cloud is approximately sorted, AFPS brings minimal performance loss while achieving a huge reduction in time cost. The method is amenable to parallel implementation and easy-to-use in current FPS-based Point cloud processing schemes.
    \item To further reduce the complexity, we propose a neighboring point distance updating scheme to reduce the number of distance updates per iteration to a constant, instead of O(N) in the original FPS. 
    \item Compared to original FPS, AFPS can achieve 22 to 30\texttimes\: speedup. With NPDU, AFPS can achieve 34\texttimes\: speedup with a slightly worse task performance of 0.8490 in instance average mean Intersection of Union (mIoU), compared to the original FPS of 0.8525 on a part-segmentation task.
\end{itemize}

The rest of the paper is organized as follows: section \ref{sec:background} introduces the background of point cloud processing and FPS. Section \ref{sec:method} presents the motivation and detail of our proposed AFPS method and NPDU heuristic. Section \ref{sec:experiments} gives extensive evaluation on performance and efficiency of proposed methods and section \ref{sec:conclusion} concludes this work.

\section{Background}
\label{sec:background}

\subsection{Point Cloud Processing}
 PointNet++ \cite{qi2017pointnet++} processes raw point cloud data directly and achieves good performance compared to traditional method based on 3D voxels or 2D views. Unlike schemes such as \cite{qi2017pointnet, ravanbakhsh2016deep} that use global max pooling of the point cloud and might miss detailed local features, PointNet++ uses set abstraction layer,  which can extract local features and gradually extend to larger global regions.
Set abstraction layer adopts sampling and grouping as the key operation for feature extraction where FPS is adopted as the sampling operation. FPS has since then been used widely in modern point cloud processing schemes \cite{qi2017pointnet++, shi2019pointrcnn, guo2021pct}. However, FPS dominates the PointNet++ inference. On GPU, for example, an RTX-2080ti GPU with 4352 CUDA cores, FPS takes about 53.7\% of the total time cost.

\subsection{Farthest Point Sampling}

\begin{figure}[htbp]
\centerline{\includegraphics[scale = 0.45]{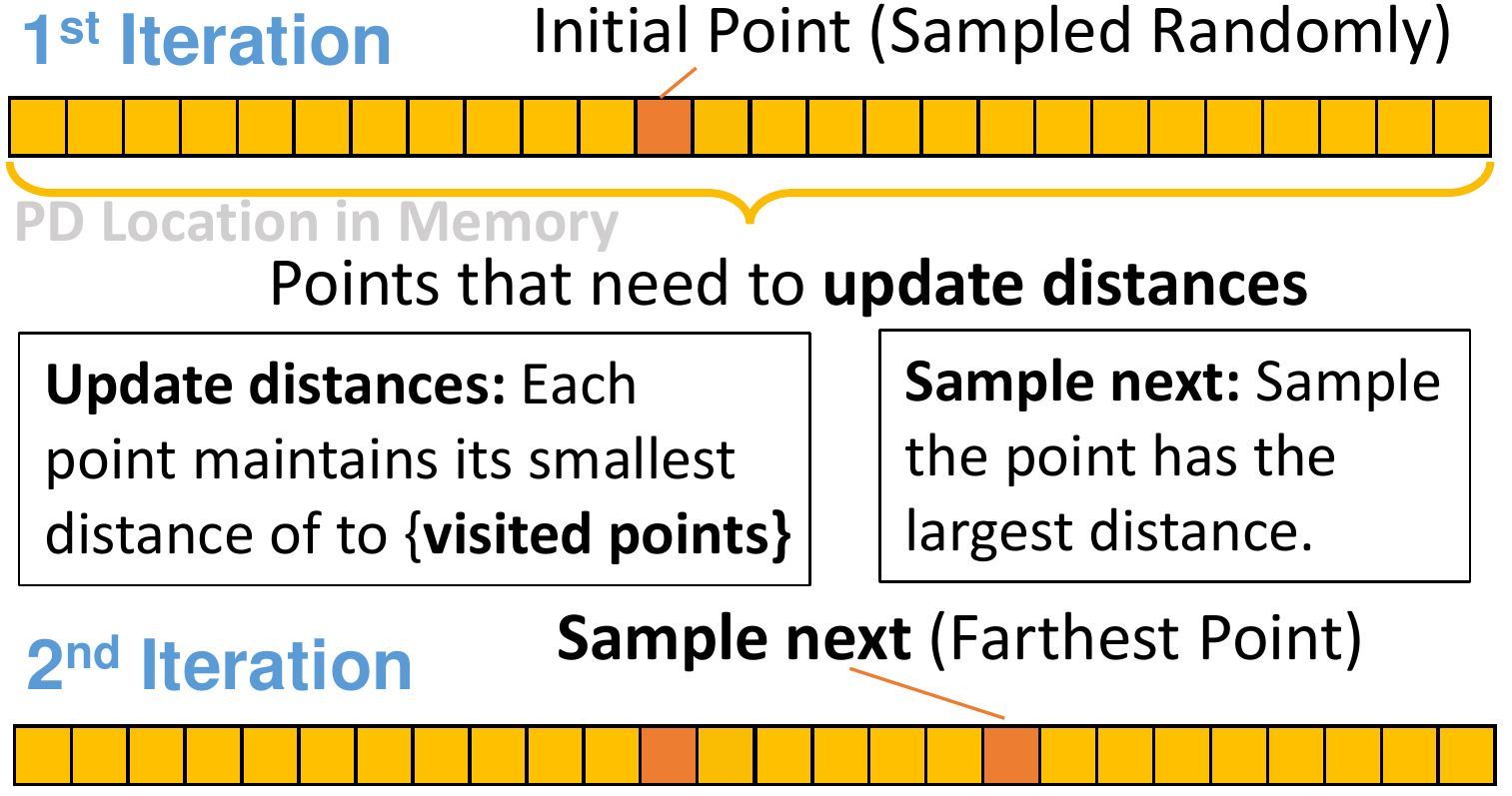}}
\caption{Visualization of FPS process.}
\label{fig:fps_how}
\end{figure}

While FPS was introduced in traditional image processing long back\cite{eldar1997farthest}, it has gained popularity with PointNet++ \cite{qi2017pointnet++}. FPS is described in Algorithm~\ref{alg:algorithm1}. FPS is a greedy algorithm that samples the point that is the farthest away to visited points in each iteration. The visualization of this process is shown in Fig.~\ref{fig:fps_how}. After the first point is sampled randomly, all other points calculate their Euclidean distance to the sampled point, and store the distance in an array. Then, the point that has the largest distance is selected as the next point to sample (line 12), and the distance array is updated if the current distance being calculated is smaller than the old distance (line 8, 9). 

We observe two important properties of FPS: 

\textbf{(P1)}: Points that are far from the newly sampled point have high priority to be sampled.

\textbf{(P2)}: In each iteration, only one point is sampled, followed by O(N) distance calculations. 

Because of \textbf{P1}, FPS is able to capture prominent point features and is robust in sampling point clouds regardless of the sparsity. However, because of \textbf{P2}, FPS has poor hardware efficiency and lacks scalability.

\begin{algorithm}[!htbp]
\SetNoFillComment
\SetAlgoLined
\SetKwInOut{Input}{Input}
\Input{Point cloud $\textbf{\textit{P}}$ of shape ($N$, $D$) denoted by number of point $N$ and point cloud dimension $D$ ($D$=3), and number of points to be sampled $n\_point$.}
\tcp{Initialize distance} 
$\textbf{\textit{Distance}}[N] = \{1e10\}$\;

$\textbf{\textit{Sampled}}[n\_point] = \{-1\}$\;

\tcp{Sampling start} 
$sample\_idx = randint(0, N)$\;

$\textbf{\textit{Sampled}}[0] =  sample\_idx$\;

\For{i $\gets$ 1 $\textbf{to}$ n\_point - 1}{
    \tcp{Update distances} 
    \For{j $\gets$ 0 $\textbf{to}$ N-1}{
        dist = geo\_dist($\textbf{P}[j, :], \textbf{P}[sample\_idx, :], D$)\;
        
        \If{dist $\leq$ \textbf{Distance}[j]}{
        \textbf{Distance}[j] = dist\;
        }
    }
    \tcp{Sample next} 
    $sample\_idx = argmax(\textbf{Distance})$\;
    
    $\textbf{Sampled}[i] =  sample\_idx$\;
    
}
\caption{Farthest Point Sampling}
\label{alg:algorithm1}
\end{algorithm}


To solve the high computation complexity of O($NC$) of FPS, the original work \cite{eldar1997farthest} as well as \cite{liu2015farthest} propose to use a tree-based data structure to reduce its complexity. However, implementing tree-based optimizations are difficult for a multi-core system such as GPU,  and implementing the original FPS is still the more favorable option \cite{openpcdet2020}.
This paper address the poor performance of FPS on a multi-core processor by proposing a heuristic version of FPS that is both light in computation and easy for parallel implementation.

There are other  alternate solutions to replace FPS and we mention them here for completeness. They include DNN-based sampling methods \cite{dovrat2019learning, lang2020samplenet} that use a sample network as a differentiable sampling and can be optimized during training, DNN-assisted sampling methods \cite{yan2020pointasnl, hu2020randla} that compensate random point sampling with a dedicated neural network module to achieve good scalability. Grid-GCN \cite{xu2020grid} adds a voxelization pre-processing step to put raw point cloud data into a regular grid before doing a scalable sampling. But this pre-processing step is non-trivial (will be shown in later experiments) and lead to poor latency in some cases.




\section{Method}
\label{sec:method}

\subsection{Dimensional Locality}

Most point cloud data is not unstructured and unsorted as previously assumed. 
In fact, point cloud data points stored in consecutive memory locations are found to be related. This is because of the nature in how they are collected. We refer to this characteristic of the data as {\it dimensional locality}. For example, point cloud data collected by 3D scanning LiDAR can be regarded as streaming data -- a fact that is exploited 
 in \cite{han2020streaming} to reduce the latency.
Another interesting case is generating 3D point cloud data from  2D-LiDAR
by attaching a step motor at the bottom of a 2D LiDAR sensor and moving the sensor vertically
\cite{fang2018real, murcia20183d}.

\begin{figure}[htbp]
\centerline{\includegraphics[scale = 0.55]{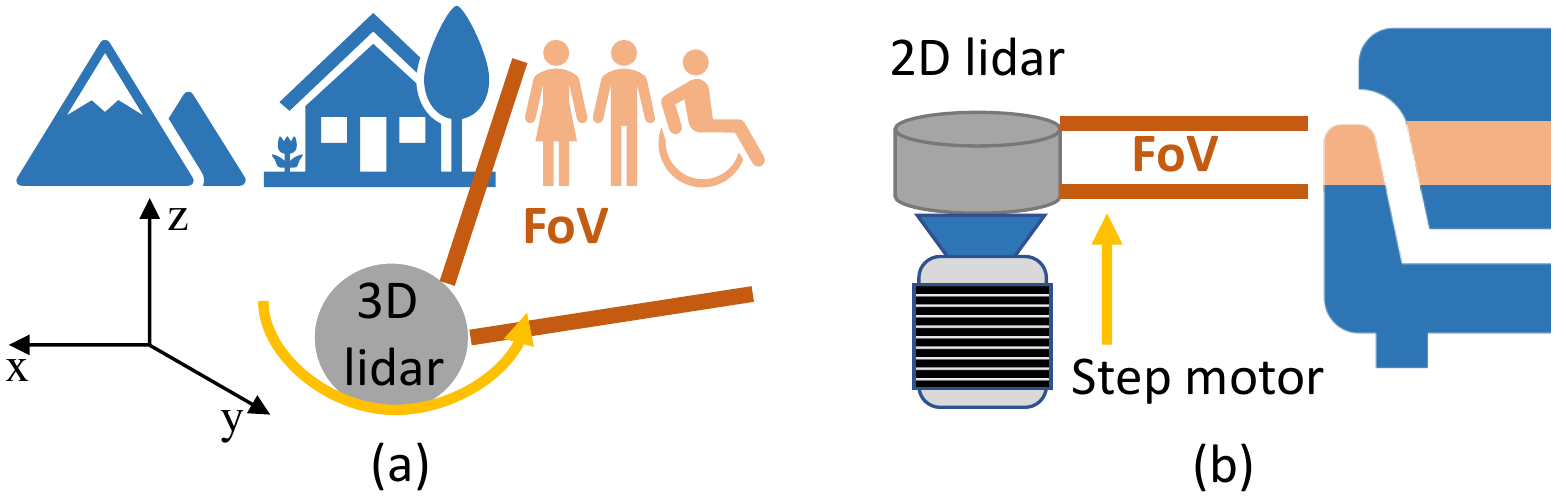}}
\caption{LiDAR point cloud data collection process. (a) Scanning 3D LiDAR, where point cloud data is approximately-sorted. (b) 2D LiDAR with stepping motor, where point cloud data is exactly-sorted. }
\label{fig:use_case}
\end{figure}

We model two data collection methods in Fig.~\ref{fig:use_case}. The scanning 3D LiDAR senses the environment by rotating around the z-axis, according to Fig.~\ref{fig:use_case} (a). Thus the point cloud can be seen as approximately sorted along the x-axis, where the bin size of the approximate sorting algorithm is dependent on the Field of View (FoV) of the sensor and proximity of the object to the sensor. The 2D LiDAR is modeled in  Fig.~\ref{fig:use_case} (b). It is equipped with a step motor and moves vertically so the point cloud data is sorted along the z-axis.

If dimensional locality is present in the point cloud,  the proposed AFPS has  minimal loss in sampling quality, as will be shown in later experiments. Dimensional locality can also be used to find approximate ``point neighbors'' for our NPDU heuristic.
For cases where point cloud data does not have locality, some form of binning can be used -- exact sorting is not required.
However, in this paper, we focus on cases that naturally have the dimensional locality and are free of sorting overhead as shown in Fig.~\ref{fig:use_case}.


\subsection{Adjustable FPS}


A direct approach to accelerate original FPS is to divide the distance calculation of all points into many cores and process them in parallel.
However, such an approach does not scale well with the number of cores. Synchronization is needed to derive global farthest point when all cores finish their distance updating in each iteration and so 
as the number of cores increases, synchronization gradually dominates the time cost, 
Our approach is quite different. We divide the original point cloud $P$ into $M$ sectors according to their storage location and perform a local FPS on each sector. Our approach reduces the total number of computations by a factor of $M$ since to sample $C$ points using AFPS, only $C/M$ iterations are needed.

However, implementing AFPS on unsorted point cloud comes with loss in sampling quality. AFPS only samples $N/M$ points separately from each sector. Therefore, if an unsorted point cloud happens to have all the ``representative points'' in the same sector, the reduced number of points (from $N$ to $N/M$) can result in missing the ``representative'' points.
But if the point cloud data has dimensional locality, the ``representative points'' are likely to be distributed more uniformly in different sectors, resulting in fewer misses. 
Thus, while this method could suffer from some small performance degradation, it results in significant reduction in number of computations. 
We set $M$ as an adjustable parameter in AFPS. 
In one extreme case, if $M$ is equal to the number of sampled points, the algorithm can finish in a single iteration and is equivalent to RPS. On the other, if $M$ is equal to one, AFPS is the same as the original FPS. We can choose the value of $M$ to achieve great latency reduction with little impact on task performance, as will be shown in the experimental results section.

\subsection{NPDU Heuristic}
From the property \textbf{P1} of the original FPS, we see that for each newly sampled point, we need to exclude points that have small distances from the points have been sampled and should encourage points that have large distances since they are better potential sampling candidates.  We want to keep this good property while limiting the number of distance updating to further reduce the computational complexity.

\begin{figure}[htbp]
\centerline{\includegraphics[scale = 0.45]{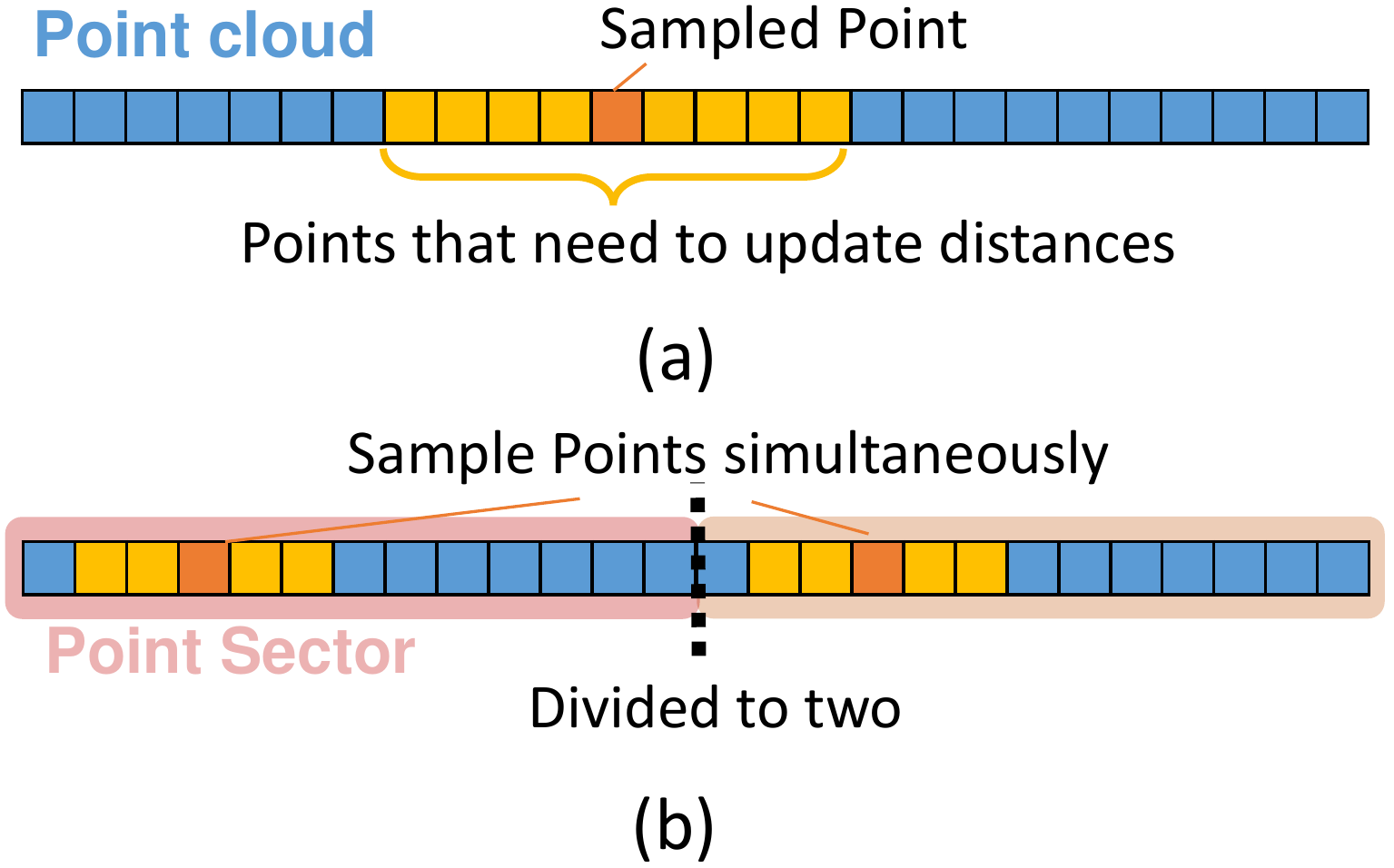}}
\caption{Illustration of NPDU heuristics. (a) Apply NPDU on original FPS with $k$ equal to 8, distance updates are done only for points that are located near the sampled points. (b) Apply NPDU on AFPS wth $M$ equal to 2 and $k$ equal to 4, point-cloud is divided into sectors, and NPDU is applied to each sector.}
\label{fig:npdu_method}
\end{figure}

Towards this goal, we propose the nearest points distance updating (NPDU) method: for a point cloud that has dimensional locality, we only update the distances of points that are stored in the nearest neighboring locations of the newly sampled point. As shown in Fig.~\ref{fig:npdu_method} (a), the horizontal bar represents a point cloud that is stored in the memory, with NPDU, only $k$ neighboring points' distances will be updated in each iteration. 
Notice that NPDU can be easily applied on top of the AFPS as shown in Fig.~\ref{fig:npdu_method} (b). It  limits the number of distance updates of each smaller-scale FPS and further reduces the computational complexity. 
NPDU heuristic preserves the nice property \textbf{P1} of the original FPS in the following manner: due to dimensional locality, it is sufficient to choose points that are approximate neighbors, that is, close in one or more dimensions. The assumption is that points that are close in one dimension are more likely to be neighbors in 3D space. Second, once distance updating of neighboring points is done, sampling of neighboring points is less likely to happen in future iterations.
By applying NPDU, we can limit distance updating in each iteration to a constant $k$ per sector.



\section{Experimental Results}
\label{sec:experiments}

\subsection{Experiment Settings}
We implement the proposed methods on top of a Pytorch repository of PointNet++ \cite{xyan2021pointnet}, which includes classification task on Modelnet40 \cite{wu20153d}, Part Segmentation task on ShapeNet \cite{chang2015shapenet} and Semantic Segmentation task on S3DIS \cite{armeni20163d}. The details of the three tasks are shown in Table.~\ref{tab:task_setting}. The input size is represented by (B, N, D), where B is the batch size, N is the number of points and D is the number of dimensions. 

We use ``task performance'' as an accuracy metric of the point cloud sampling method. For the evaluations presented here, we apply sampling methods in a pre-trained PointNet++ model, and perform testing on down-streaming tasks.
For instance, we use class accuracy for object classification task and sematic segmentation task, and instance average mIoU (mean Intersection of Union) for part segmentation task \footnote{ For detailed definition of each metric, please refer to \href{https://github.com/yanx27/Pointnet\_Pointnet2\_pytorch}{\tocheck{Pointnet\_Pointnet2\_pytorch}}}.
We take the average of 10 evaluations for each data point to eliminate randomness.

\begin{table}[htbp]
\caption{Detail of three PointNet++ tasks (batch size = 24).}
\centering
\resizebox{1.0\linewidth}{!}{
\begin{tabular}{lcc}
 \toprule
 Task & Input Size & Evaluation Metric\\
 \midrule
 Object Classification & (24, 1024,3) & Class Accuracy \\
 Part Segmentation & (24, 2048,3) & Instance Average mIoU\\
 Semantic Segmentation & (24, 4096,3) & Class Accuracy\\
 \bottomrule
\end{tabular}
\label{tab:task_setting}
}
\end{table}

For evaluation, we use Transmuter~\cite{transmuter} (TM) as the testbed. TM is a multi-core architecture that optimizes both programmability and efficiency. We use a configuration with 4 tiles, where each tile consists of one LCP (local control processor) and 16 Arm cores. Cache memory design comprises shared cache of 4KB L1 cache bank per core (shared across cores) and 4KB L2 cache bank per tile (shared across tiles). We use its Gem-5 implementation for cycle-accurate simulation to get latency and cache statistics.


\subsection{AFPS evaluation}

\begin{figure}[htbp]
\centerline{\includegraphics[scale = 0.8]{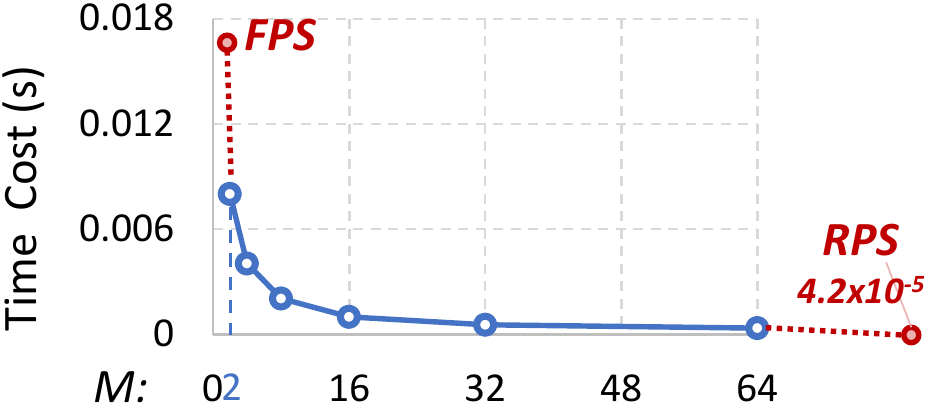}}
\caption{Time cost of the proposed AFPS on part segmentation task measured on TM testbed. for different values of $M$. The input size is (24, 2048, 3) and the sample size is N=512.}
\label{fig:parallel_M}
\end{figure}

We test the speedup of the proposed AFPS sampling method on the part segmentation task. As is shown in Fig.~\ref{fig:parallel_M}, compared to original FPS, AFPS reduces time cost significantly with increasing $M$. It saturates when $M$ increases above 16 because GPE to LCP communication gradually dominates the time cost. At $M = 32$, the time cost is $5.7\times10^{-4}$ second, a 29.8\texttimes\: speedup compared to the original FPS which takes $0.017$ second. 
Thus, setting $M = 32$ (or even lower) for this TM configuration is sufficient for addressing the bottleneck issue for this application.
Note that choosing a large $M$ provides marginal improvement on FPS performance and does not translate to the overall latency reduction as other components become the bottleneck.

Next, we investigate the task performance of the proposed AFPS on the part segmentation task, where we compare its performance on unsorted, approximately sorted, and exactly sorted point clouds. 
The results are shown in Table.~\ref{tab:afps_sample_perf}. The baseline original FPS achieves an average mIoU of 0.8522 on the part segmentation task. 
If the point cloud is unsorted, applying AFPS suffers a large performance drop: When $M$ is 32, it only achieves mIoU of 0.8447. 
However, if the point cloud is approximately sorted along dimension $x$, the performance drop is much smaller. For approximately sorted data, we set the bin size\footnote{every point in a bin with a larger index is greater than points in a bin with a smaller index, but points in a bin are unsorted.} at 128 points. For $M$ is 2 or 8, AFPS on approximately-sorted point cloud has almost no drop in task performance compared to the baseline.
We also have a similar observation on a point cloud that is exactly sorted. For 2D LiDAR with stepping motor case where the point cloud is exactly sorted along $z$ axis, the performance even increases to 0.8527 mIoU for $M$ of 32, which is even higher than that of the original FPS.

\begin{table}[htbp]
\vspace{-5pt}
\caption{Task performance of proposed AFPS on point clouds with different characteristics. \textbf{(Baseline: 0.8522)}}
\centering
\resizebox{0.85\linewidth}{!}{
\begin{tabular}{lccc}
 \toprule
 \multirow{2}{*}{\textbf{Characteristics}} & \multicolumn{3}{c}{\textbf{Task Performance (mIoU)}}\\
  & $M=2$ & $M=8$ & $M=32$\\
 \midrule
 unsorted & 0.8503 & 0.8468 & 0.8447 \\
 approx.-sorted (x) & 0.8526 & 0.8518 & 0.8488 \\
 exactly-sorted (x) & 0.8519 & 0.8515 & 0.8489 \\
 exactly-sorted (z) & 0.8493 & 0.8523 & 0.8527 \\
 \bottomrule
\end{tabular}
}
\label{tab:afps_sample_perf}
\vspace{-5pt}
\end{table}

\subsection{NPDU on original FPS}
\begin{figure}[htbp]
\centerline{\includegraphics[scale = 0.76]{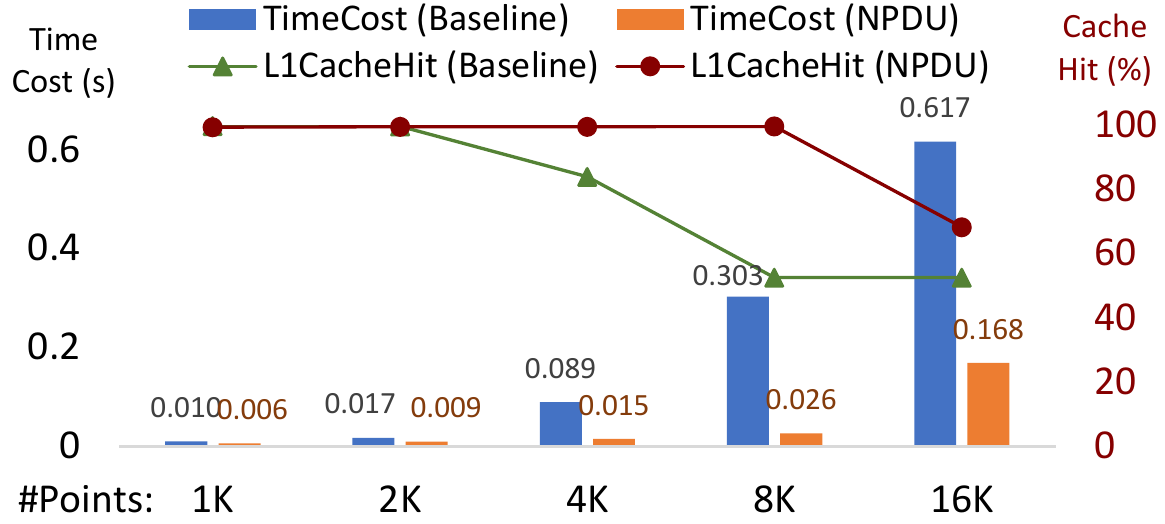}}
\caption{Time cost of FPS (baseline) and NPDU for varying number of points of input sizes while keeping sample size to be 512. NPDU has good performance even for larger input data sizes.}
\label{fig:sequential_scale}
\end{figure}

Next, we test the proposed NPDU heuristic in terms of hardware efficiency. The number of points $N$ of the input point cloud is varied from 1K to 16K and $k$ is set equal to 8. The TM time cost and cache hit rate performance is shown in Fig.~\ref{fig:sequential_scale}. 
When the point cloud size is small at 1K/2K, applying NPDU achieves a 2\texttimes\: speedup but when the input size increases to 8K, a peak speedup of around 11x can be achieved. The reason of the large gain is that while the cache hit rate of FPS drops significantly, the cache hit rate of NPDU remains the same. 
However, when the point cloud size data increases to 16K, the difference in the cache hit rates reduces and the time cost difference between the two methods is not as spectacular.

\begin{figure}[htbp]
\centerline{\includegraphics[scale = 0.76]{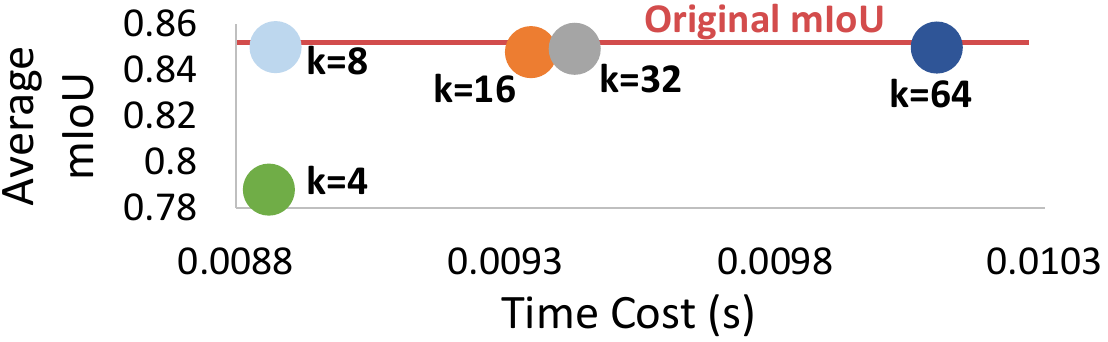}}
\caption{Task performance (mIoU) and time cost of part segmentation task as a function of
$k$. The input size is 2048 and the sample size is 512.}
\label{fig:sequential_scale2}
\end{figure}

We investigate the task performance and efficiency tradeoff of NPDU on the part segmentation task for different values of $k$ (4 to 64). As shown in Fig.~\ref{fig:sequential_scale2}, choosing a small $k$ brings the lowest time cost but results in a noticeable degradation in task performance, compared to original FPS that is represented by the red line.
We observe choosing $k$ equal to 8 can achieve optimal mIoU performance with a lower time cost compared to larger $k$. 
However, the optimal point differs from case to case, especially when using NPDU with AFPS.



\subsection{NPDU on AFPS}
\begin{figure}[htbp]
\centerline{\includegraphics[scale = 0.74]{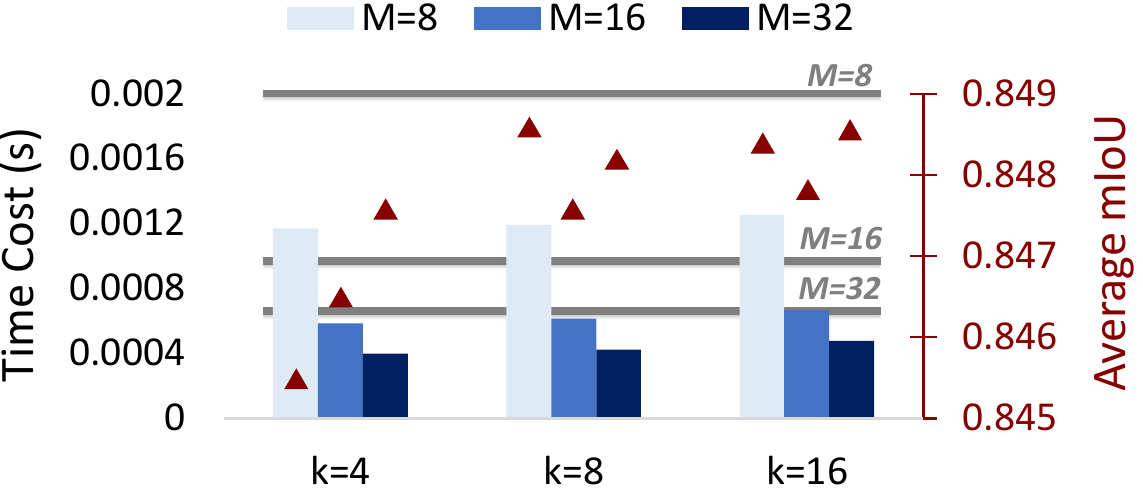}}
\caption{Time cost on TM and average mIoU for different values of $M$ and $k$ of NPDU AFPS. The input size 2048 and the sample size is 512. The three gray rails are AFPS w/o NPDU.}
\label{fig:NPDU_M}
\end{figure}

Next, we study a combination of applying NPDU on AFPS, which further reduces the time cost. In Fig.~\ref{fig:NPDU_M}, the time cost of the baseline AFPS with different values of $M$ are shown as the background gray grids. We can see a clear difference in time cost between different values of $M$. The time cost of NPDU with AFPS is shown by the blue bars.  We see that NPDU with AFPS achieves 30\% to 50\% time cost reduction compared to the AFPS baseline (lower than gray grids). 
We also observe that for small $M$, using a small $k$ can impact the average mIoU task performance a lot. While other cases do not present obvious trend. To achieve the optimal tradeoff point, $k$ needs to be tuned differently for different choices of $M$ through experimentation. For the part segmentation task, the optimal $k$ is 8 for $M=8$ and the optimal $k$ is 16 for $M=16$.


We also observe that AFPS in conjunction with NPDU is suitable for multi-core hardware such as TM.
For low $k$, when the number of cores (of TM) increases, NPDU has lower core utilization which results in speedup degradation. For instance, when the number of cores equals to 16, the speedup over a single core is only 13.72\texttimes\:(instead of 16\texttimes\:), as shown in Table.~\ref{tab:core_scale}.
This is because the workload of only $k$ distance calculations per iteration is small for distribution among 16 cores. When combined with AFPS, the total workload becomes $k \times M$, and the speedup is steady at around 2\texttimes\: when number of cores doubles.

\begin{table}[htbp]
\caption{Speedup over single core}
\centering
\resizebox{0.95\linewidth}{!}{
\begin{tabular}{lcccc}
 \toprule
 \multirow{2}{*}{\textbf{Method}} & \multicolumn{4}{c}{\textbf{Number of cores}}\\
  & 2 & 4 & 8 & 16\\
 \midrule
 NPDU FPS (k=8)& 2.19x & 4.27x & 8.72x & \textbf{13.72x}\\
 NPDU AFPS (M=16, k=4) & 2.02x & 4.28x & 8.99x & 17.17x\\
 NPDU AFPS (M=32, k=16)& 1.96x & 4.01x & 8.79x & 16.78x\\
 \bottomrule
\end{tabular}
}
\label{tab:core_scale}
\end{table}

The proposed NPDU AFPS can easily scale up to a larger problem. As shown in Table.~\ref{tab:NPDUAPFS_scale}, NPDU AFPS with $M$ set to 32 achieves much better scalability. Using all 16 cores, AFPS can sample a 32K large point cloud in just 0.0022s and achieve a massive speedup of 280\texttimes\: compared to the original FPS.

\begin{table}[htbp]
\caption{Time Cost (s) for different point cloud input sizes}
\centering
\begin{tabular}{lccccc}
 \toprule
 \multirow{2}{*}{\textbf{Method}} & \multicolumn{5}{c}{\textbf{Point Cloud Input Sizes}}\\
  & 2k & 4k & 8k & 16k & 32k\\
 \midrule
 FPS& 0.0170 & 0.0892 & 0.3028 & 0.6172 & 1.2211\\
 NPDU AFPS (M=32)& 0.0005 & 0.0007 & 0.0011 & 0.0022 & 0.0044\\
 \bottomrule
\end{tabular}
\label{tab:NPDUAPFS_scale}
\end{table}




\subsection{Comparisons}
\begin{figure}[htbp]
\centerline{\includegraphics[scale = 0.78]{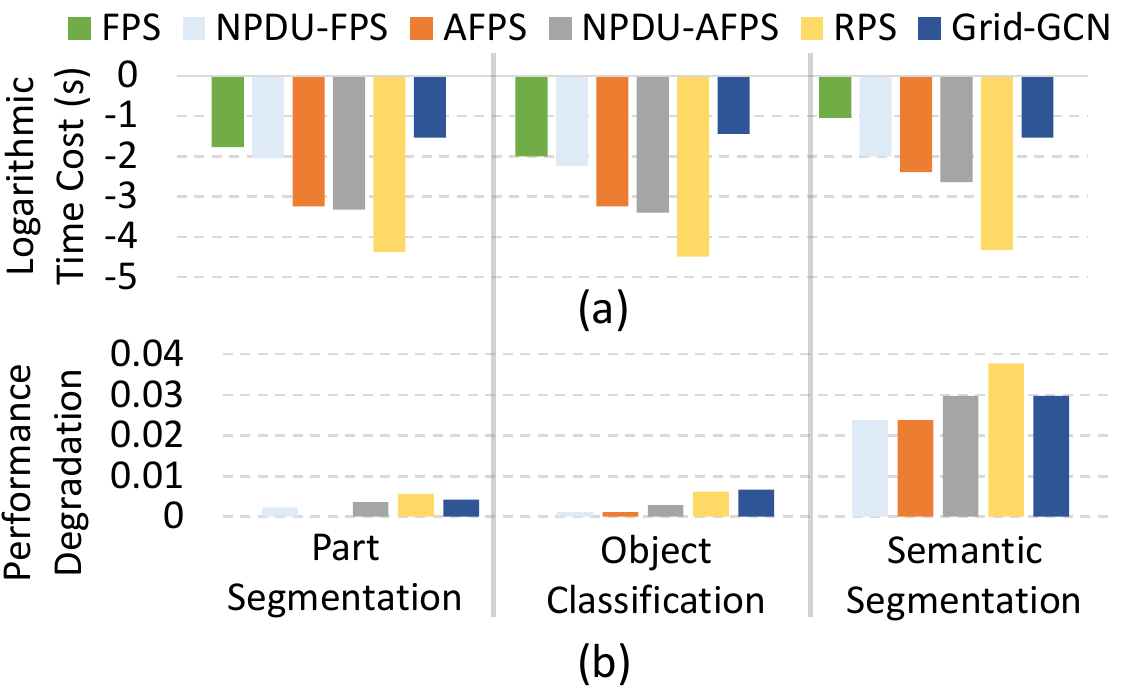}}
\caption{Comparison of different sampling methods. (a) logarithm (base 10) of time cost (base 10). (b) Performance degradation compared to original FPS.}
\label{fig:comparison}
\end{figure}

We compare both the time cost and task performance of other sampling methods with our proposed method. Comparison with Grid-GCN is done by implementing both voxelization that partitions point cloud into 3D grids with 40 grids in each dimension, and the random voxel sampling method used in the original paper \cite{xu2020grid} in C++ on the TM platform. We also include RPS and original FPS in our evaluations. For task performance, we include all three tasks as described in Table~\ref{tab:task_setting}.
Fig.~\ref{fig:comparison} shows time cost in logarithmic scale in (a) and performance degradation compared to original FPS in (b).
For time cost comparison, we can see that RPS achieves the lowest time cost while FPS and Grid-GCN have similar highest time cost. The gray bar that represents our proposed NPDU AFPS achieves the second-lowest time cost. Grid-GCN has poor time cost performance because of the time-consuming voxelization step. It costs even more time than the original FPS for part segementation and object classification because of smaller number of points.
For task performance, RPS has the highest degradation, whereas proposed NPDU FPS and AFPS achieve the smallest degradation. NPDU AFPS is slightly worse than AFPS, but it is consistently better than Grid-GCN or RPS in all three tasks.

\section{Conclusion}
\label{sec:conclusion}
In this work, we propose AFPS method that separates point cloud into multiple small sectors along a sorted dimension, and NPDU heuristic that limits the number of distance updates. We exploit the approximately-sorted point cloud pattern in the data collection step and use it to our advantage to compensate for the task performance degradation. 
Experimental results show the improvement on latency and scalability over the original FPS method. 
NPDU in combination with AFPS can achieve a 34\texttimes\: speedup on a 2K point cloud and 280\texttimes\: speedup on a 32K point cloud, compared to the original FPS. For task performance, the proposed NPDU AFPS has consistently better performance than RPS or Grid-GCN. While this work focused on demonstration of AFPS with NPDU on a multi-core hardware, we plan to implement it in a GPU in the near future.




\bibliographystyle{IEEEtran}
\bibliography{IEEEtran}
\end{document}